\documentclass{article} %
\usepackage{iclr2023_conference}

\usepackage{booktabs,arydshln}
\makeatletter
\def\adl@drawiv#1#2#3{%
        \hskip.5\tabcolsep
        \xleaders#3{#2.5\@tempdimb #1{1}#2.5\@tempdimb}%
                #2\z@ plus1fil minus1fil\relax
        \hskip.5\tabcolsep}
\newcommand{\cdashlinelr}[1]{%
  \noalign{\vskip\aboverulesep
           \global\let\@dashdrawstore\adl@draw
           \global\let\adl@draw\adl@drawiv}
  \cdashline{#1}
  \noalign{\global\let\adl@draw\@dashdrawstore
           \vskip\belowrulesep}}
\makeatother

\usepackage{times}

\usepackage{amsmath,amsfonts,bm}

\def\eqref#1{equation~\ref{#1}}

\def\1{\bm{1}}

\DeclareMathAlphabet{\mathsfit}{\encodingdefault}{\sfdefault}{m}{sl}
\SetMathAlphabet{\mathsfit}{bold}{\encodingdefault}{\sfdefault}{bx}{n}

\usepackage[]{xcolor}
\usepackage{array}
\usepackage{times}
\usepackage{latexsym}
\usepackage[T1]{fontenc}
\usepackage[utf8]{inputenc}
\usepackage{microtype}
\usepackage[utf8]{inputenc} %
\usepackage[T1]{fontenc}    %
\usepackage{url}            %
\usepackage{booktabs}       %
\usepackage{amsfonts}       %
\usepackage{nicefrac}       %
\usepackage{microtype}      %
\usepackage{changepage}
\usepackage{xargs}          %
\usepackage{wrapfig,lipsum,booktabs}
\usepackage{enumitem}
\usepackage{longtable}
\usepackage{makecell}
\usepackage{graphicx}
\usepackage{subcaption}
\usepackage{pifont}
\definecolor{mydarkblue}{rgb}{0,0.08,0.45}
\usepackage[colorlinks,citecolor=mydarkblue,urlcolor=mydarkblue,linkcolor=mydarkblue]{hyperref}

\usepackage{multirow, makecell}
\usepackage{tabularx}
\usepackage{xspace,mfirstuc,tabulary}

\usepackage{pgfplots}
\usetikzlibrary{pgfplots.groupplots}
\pgfplotsset{compat=1.3}
\usepackage{tikz}
\usetikzlibrary{patterns}

\usepackage[most]{tcolorbox}

\usepackage{changepage}
\usepackage[scaled=1]{couriers}

\usepackage{todonotes}
\makeatletter
\newcommand*\iftodonotes{\if@todonotes@disabled\expandafter\@secondoftwo\else\expandafter\@firstoftwo\fi} 
\makeatother

\newcommand{\sbt}{\,\begin{picture}(-1,1)(-1,-3)\circle*{3}\end{picture}\ }
\newcommand{\mgsm}{{MGSM}\xspace}
\newcommand{\xcopa}{{XCOPA}\xspace}
\newcommand{\xlwic}{{XL-WiC}\xspace}

\newcommand{\openai}[3]{$\mathtt{#1}$-$\mathtt{#2}$-$\mathtt{#3}$\xspace}

\newcommand{\PaLM}{PaLM\xspace}
\newcommand{\InstructGPT}{GPT-3\xspace}

\newcommand{\interalia}[1]{\citep[\textit{inter alia}]{#1}}
\newcommand{\smallbullet}[0]{\sbt\ \ }
\newcommand{\direct}{\textsc{Direct}\xspace}
\newcommand{\nativecot}{\textsc{Native-CoT}\xspace}
\newcommand{\encot}{\textsc{EN-CoT}\xspace}
\newcommand{\translateen}{\textsc{Translate-EN}\xspace}
\newcommand{\nativeexemplar}{\textsc{Native-Exemplars}\xspace}
\newcommand{\enexemplar}{\textsc{English-Exemplars}\xspace}
\newcommand{\multilingualexemplar}{\textsc{Multilingual-Exemplars}\xspace}

\definecolor{battleshipgrey}{rgb}{0.3, 0.3, 0.3}
\definecolor{brilliantrose}{rgb}{1.0, 0.33, 0.64}
\definecolor{americanrose}{rgb}{1.0, 0.01, 0.24}
\definecolor{jweigreen}{rgb}{0,0.45,0.24}
\definecolor{bluegray}{rgb}{0.1, 0.1, 0.4}
\definecolor{ao(english)}{rgb}{0.0, 0.5, 0.0}
\definecolor{blanchedalmond}{rgb}{1.0, 0.92, 0.8}
\definecolor{atomictangerine}{rgb}{1.0, 0.6, 0.4}
\definecolor{chocolate(web)}{rgb}{0.82, 0.41, 0.12}
\definecolor{bananayellow}{rgb}{1.0, 0.88, 0.21}
\definecolor{goldenbrown}{rgb}{0.6, 0.4, 0.08}
\definecolor{aliceblue}{rgb}{0.94, 0.97, 1.0}
\definecolor{beige}{rgb}{0.96, 0.96, 0.86}
\definecolor{babyblue}{rgb}{0.54, 0.81, 0.94}
\definecolor{camel}{rgb}{0.76, 0.6, 0.42}
\definecolor{cinnamon}{rgb}{0.82, 0.41, 0.12}
\definecolor{deepskyblue}{rgb}{0.0, 0.75, 1.0}
\definecolor{frenchblue}{rgb}{0.0, 0.45, 0.73}
\definecolor{classicrose}{rgb}{0.98, 0.8, 0.91}
\definecolor{frenchrose}{rgb}{0.96, 0.29, 0.54}
\definecolor{frenchlilac}{rgb}{0.53, 0.38, 0.56}
\definecolor{frenchbeige}{rgb}{0.65, 0.48, 0.36}

\newcommand{\pl}[2][]{}
\newcommand{\plresolved}[2][]{}

\usepackage[capitalize,noabbrev]{cleveref}
\crefname{section}{\S}{\S\S}
\Crefname{section}{\S}{\S\S}
\crefname{table}{Table}{Tables}
\crefname{figure}{Figure}{Figures}
\crefname{algorithm}{Algorithm}{}
\crefname{equation}{eq.}{}
\crefname{appendix}{Appendix}{}
\crefformat{section}{\S#2#1#3}
\usepackage{float}

\title{Language Models are \\Multilingual Chain-of-Thought Reasoners}

\author{
\hspace{-1.8mm} Freda Shi$^{1, 2,}$\thanks{Equal contribution. Work done during internship at Google Research.} \hspace{5mm} Mirac Suzgun$^{1, 3, *}$ \hspace{5mm} Markus Freitag$^1$ \hspace{5mm} Xuezhi Wang$^1$ 
\And Suraj Srivats$^4$ \hspace{5mm} Soroush Vosoughi$^4$ \hspace{5mm} Hyung Won Chung$^1$ \hspace{5mm} Yi Tay$^1$ 
\And  Sebastian Ruder$^1$  \hspace{5mm} Denny Zhou$^1$  \hspace{5mm} Dipanjan Das$^1$  \hspace{5mm} Jason Wei$^1$ \\[5mm]
$^1$Google Research \hspace{5mm}
$^2$Toyota Technological Institute at Chicago  \hspace{5mm} \\
$^3$Stanford University  \hspace{5mm}
$^4$Dartmouth College
}

\iclrfinalcopy %
\begin{document}

\maketitle
\begin{abstract}
We evaluate the reasoning abilities of large language models in multilingual settings.
We introduce the Multilingual Grade School Math (\mgsm) benchmark, by manually translating 250 grade-school math problems from the GSM8K dataset \citep{cobbe2021training} into \textit{ten} typologically diverse languages.
We find that the ability to solve \mgsm problems via chain-of-thought prompting emerges with increasing model scale, and that models have strikingly strong multilingual reasoning abilities, even in underrepresented languages such as Bengali and Swahili.
Finally, we show that the multilingual reasoning abilities of language models extend to other tasks such as commonsense reasoning and word-in-context semantic judgment.
The MGSM benchmark is publicly available at \url{https://github.com/google-research/url-nlp}.
\end{abstract}

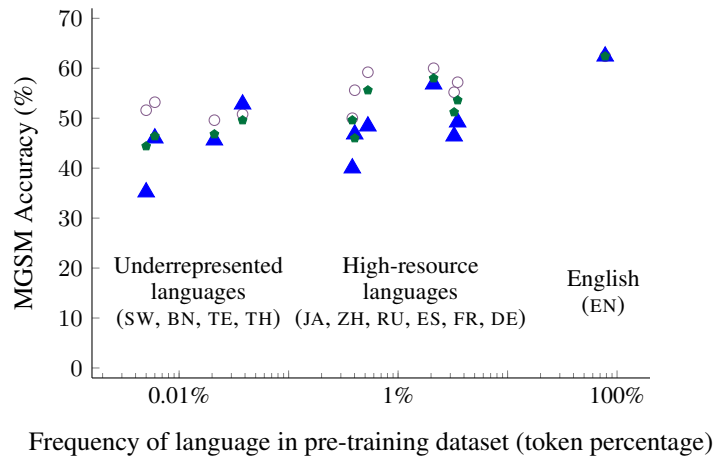
\begin{figure}[h]
\begin{centering}
    \begin{tikzpicture}
        \pgfplotsset{footnotesize,samples=10}
        \begin{groupplot}[
            group style = {group size = 2 by 1, horizontal sep = 30pt},
            width = 9cm, 
            height = 6.5cm]
            \nextgroupplot[
                align = center,
                legend style={at={(0.5,1.05)},anchor=south,draw=none},
                legend cell align={left},
                xmode=log,
                xmin=0.000016, xmax=2,
                ymin=-2, ymax=72,
                xtick={0.00001, 0.0001, 0.001, 0.01, 0.1, 1},
                xticklabels={, 0.01\%, , 1\%, , 100\%},
                axis x line*=bottom,
                axis y line*=left,
                xlabel={Frequency of language in pre-training dataset (token percentage)},
                ylabel={\mgsm{} Accuracy (\%)},
                ytick={0, 10, 20, 30, 40, 50, 60, 70},
                grid style=dashed,
                x label style={at={(axis description cs:0.5,-0.12)},anchor=north,font=\normalsize},
                y label style={at={(axis description cs:-0.08,0.5)},anchor=south,font=\normalsize},
                xtick pos=bottom,
                ytick pos=left,
                ]
                \addplot[
                    color=frenchlilac,
                    mark=o,
                    mark size=2pt,
                    only marks,
                    ]
                    coordinates {
                    (0.779, 62.4) %
                    (0.03501, 57.2) %
                    (0.0325, 55.2) %
                    (0.02112, 60.0) %
                    (0.0053, 59.2) %
                    (0.00402, 55.6) %
                    (0.00382, 50.0) %
                    (0.00038, 50.8) %
                    (0.00021, 49.6) %
                    (0.00006, 53.2) %
                    (0.00005, 51.6) %
                    };
                    \addlegendentry{\hspace{1mm}Translate to English with Google Translate and solve with English intermediate steps}
                \addplot[
                    color=blue,
                    mark=triangle*,
                    mark size=3.6pt,
                    only marks,
                    ]
                    coordinates {
                    (0.779, 62.4) %
                    (0.03501, 49.2) %
                    (0.0325, 46.4) %
                    (0.02112, 56.8) %
                    (0.0053, 48.4) %
                    (0.00402, 46.8) %
                    (0.00382, 40.0) %
                    (0.00038, 52.8) %
                    (0.00021, 45.6) %
                    (0.00006, 46.0) %
                    (0.00005, 35.2) %
                    };
                    \addlegendentry{\hspace{1mm}Intermediate reasoning steps in the language of the question}
                \addplot[
                    color=jweigreen,
                    mark=pentagon*,
                    mark size=1.7pt,
                    only marks,
                    ]
                    coordinates {
                    (0.779, 62.4) %
                    (0.03501, 53.6) %
                    (0.0325, 51.2) %
                    (0.02112, 58.0) %
                    (0.0053, 55.6) %
                    (0.00402, 46.0) %
                    (0.00382, 49.6) %
                    (0.00038, 49.6) %
                    (0.00021, 46.8) %
                    (0.00006, 46.4) %
                    (0.00005, 44.4) %
                    };
                    \addlegendentry{\hspace{1mm}Intermediate reasoning steps in English}
                    \node[font=\small] at (axis cs: 0.00015,15) {Underrepresented \\ languages \\ (\textsc{sw}, \textsc{bn}, \textsc{te}, \textsc{th}) };
                    \node[font=\small] at (axis cs: 0.013,15) {High-resource \\ languages \\ (\textsc{ja}, \textsc{zh}, \textsc{ru}, \textsc{es}, \textsc{fr}, \textsc{de})};
                    \node[font=\small] at (axis cs: 0.75,15) {English \\ (\textsc{en})};
        \end{groupplot}
    \end{tikzpicture}
    \caption{Correlation between language frequency and \mgsm accuracy for \PaLM-540B. The accuracy is surprisingly high, even for underrepresented languages like Swahili (\textsc{sw}) and Bengali (\textsc{bn}), which account for less than 0.01\% of the pre-training dataset.} 
    \label{fig:freq_acc}
    \end{centering}
\end{figure}
\section{Introduction}

Recent work has shown that presenting explicit reasoning steps (i.e., chains of thought; \textsc{CoT}) in English elicits multi-step reasoning abilities of large language models such as GPT-3 and PaLM \interalia{brown2020language,chowdhery2022palm,wei2022chain}.
Pretrained multilingual language models have also achieved impressive performance on various NLP tasks across typologically distinct languages \interalia{conneau-etal-2020-unsupervised,xue-etal-2021-mt5,chowdhery2022palm,clark-etal-2020-tydi,hu2020xtreme,ruder-etal-2021-xtreme}. 
Tasks in existing multilingual benchmarks usually require only simple reasoning steps, and so it is still unclear how well language models perform on tasks that require more complex reasoning in a multilingual setting. 

In this work, we introduce the \textbf{\mgsm} benchmark to bridge the gap between the progress on English-based chain-of-thought reasoning and multilingual NLP. We extend a subset of the English-language GSM8K dataset \citep{cobbe2021training} to ten typologically diverse languages via manual translation of problems into target languages. To the best of our knowledge, this is the first multilingual benchmark to evaluate the arithmetic reasoning abilities of language models.

We evaluate two large language models, \InstructGPT \citep{brown2020language,ouyang2022training} and \PaLM \citep{chowdhery2022palm},
on this benchmark. While both models solve less than 20\% of problems with standard prompting, the 540-billion-parameter \PaLM model in particular shows exceptional multilingual reasoning abilities with intermediate reasoning steps (\cref{fig:freq_acc}), solving more than 40\% of the problems in any investigated language, including underrepresented languages such as Bengali and Swahili. In our best setting, \PaLM achieves an average solve rate of 55\% across languages.
We find that intermediate reasoning steps in English consistently lead to competitive or better results than those written in the native language of the question, suggesting that English chain-of-thought prompting may be a useful baseline for future multilingual reasoning work.

We further demonstrate that the multilingual reasoning abilities of pretrained models extend to common-sense reasoning \citep{ponti-etal-2020-xcopa} and word-in-context semantic judgment \citep{raganato-etal-2020-xl}.
By presenting the models with few-shot examples in different languages, \PaLM
sets a new state-of-the-art performance ($89.9\%$) on \xcopa \citep{ponti-etal-2020-xcopa}, outperforming the prior approaches that require thousands of training examples.

\section{The \mgsm Benchmark}

\label{sec:dataset-mgsm}
In this section, we describe the collection process of Multilingual Grade School Math (\mgsm), to our knowledge the first multilingual arithmetic reasoning benchmark.

\begin{wrapfigure}{r}{0.41\textwidth}
    \centering
    \includegraphics[width=0.41\textwidth]{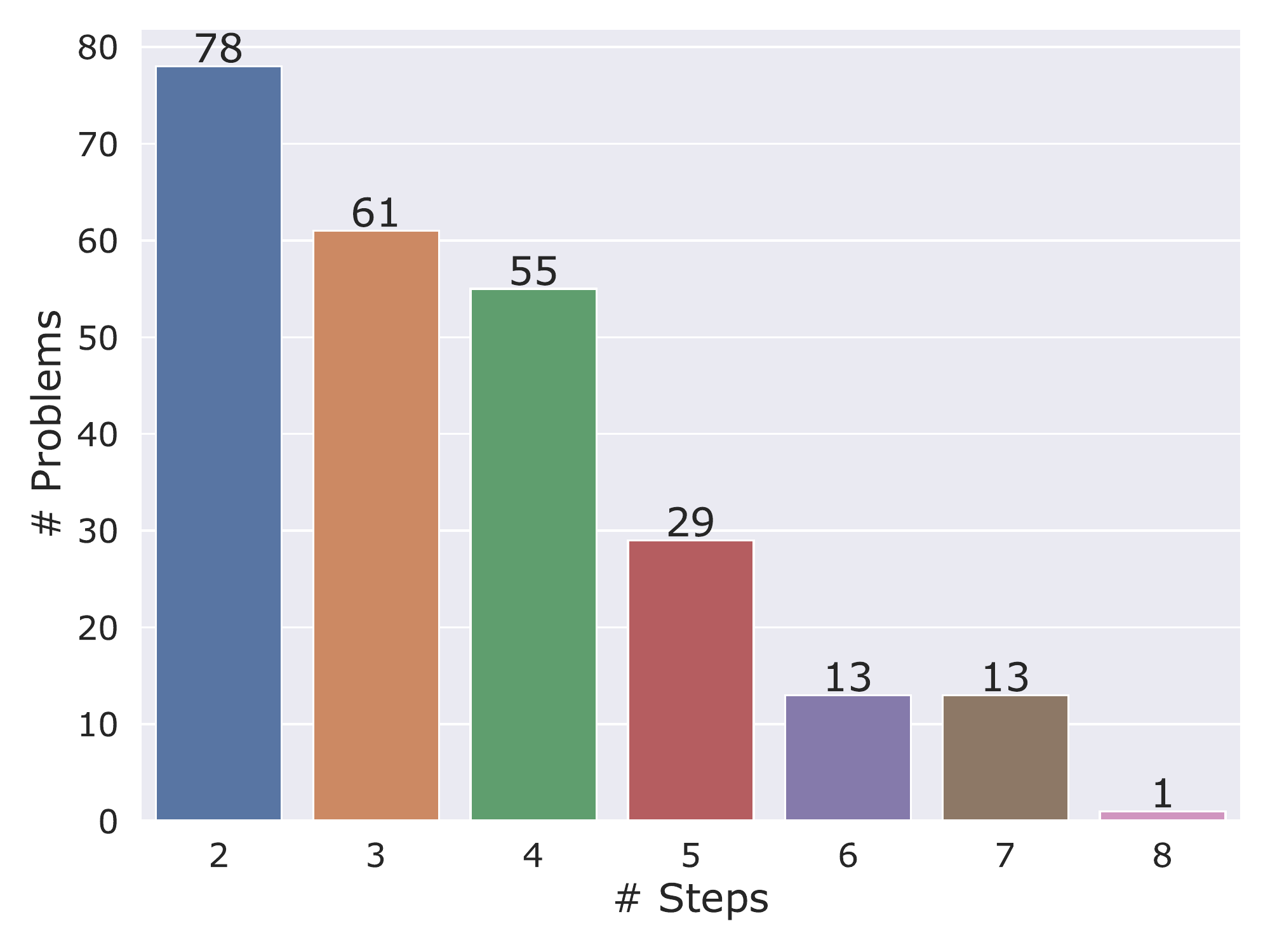}
    \caption{\mgsm problem distribution with respect to the number of reasoning steps in the standard solution.}
    \vspace{-5mm}
    \label{fig:mgsm-stats}
\end{wrapfigure}

\paragraph{Source data.}
We used GSM8K \citep{cobbe2021training}, an English-language human-annotated grade-school math problem dataset, as the base data source.
For \mgsm, we took the first 250 examples from the GSM8K official test example list.
Each problem requires two to eight steps to solve according to the official solution (Figure~\ref{fig:mgsm-stats}).
The answer for each question in GSM8K was written as an Arabic numeral, which we kept consistent across all languages to facilitate cross-lingual prediction.\footnote{Certain scripts such as Devanagari employ different numerals. We restrict the data to Arabic numerals for consistency but future work may investigate cross-lingual numeracy by mapping Arabic numerals to those of the corresponding script \citep[see][]{spithourakis-riedel-2018-numeracy}.}

\paragraph{Target language selection.}
We selected a typologically diverse set of ten languages other than English (\textsc{en}), spanning eight language families and different levels of representation in standard pretraining datasets such as mC4 \citep{xue-etal-2021-mt5}: Bengali (\textsc{bn}), Chinese (\textsc{zh}), French (\textsc{fr}), German (\textsc{de}), Japanese (\textsc{ja}), Russian (\textsc{ru}), Spanish (\textsc{es}), Swahili (\textsc{sw}), Telugu (\textsc{te}), and Thai (\textsc{th}).

\paragraph{Manual translation process.}
We enlisted the help of paid professional translators (two for Chinese and German, three for Russian, five for Thai, one for each remaining target language) for the manual translation of the 250 selected English-language examples from GSM8K.
All translators involved were native speakers of the target language and had at least two years of professional experience in translating between English and the target language.
All translators had signed a machine translation (MT) non-usage declaration before they started to work.
To verify the quality of the human translations, the vendor sent a random subset of translations to an additional translator to verify the quality, and checked for $n$-gram overlap with popular MT providers to ensure that no machine translation toolkit has been used. 
We employ the translation results as gold standard translations.

\section{Multilingual Chain-of-Thought Prompting}
\label{sec:approaches}
\begin{table}[t]
    \centering \small
    \begin{tabular}{p{0.19\textwidth}p{0.7\textwidth}}
    \toprule
    \bf Original Question & \underline{Frage: Roger hat 5 Tennisbälle. Er kauft noch 2 Dosen Tennisbälle. In jeder } 
    \underline{Dose sind 3 Tennisbälle. Wie viele Tennisbälle hat er jetzt?} \\
    \midrule
    \direct & \underline{Antwort}: 11 \\
    \midrule
    \nativecot & \underline{Schritt-für-Schritt-Antwort: Roger begann mit 5 Bällen. 2 Dosen von jeweils 3 }
    \underline{Tennisbällen macht 6 Tennisbälle. 5 + 6 = 11. Die Antwort ist} 11. \\
    \midrule
    \encot & Step-by-Step Answer: Roger started with 5 balls. 2 cans of 3 tennis balls each is 6 tennis balls. 5 + 6 = 11. The answer is 11. \\
    \midrule \midrule
    \bf Translated\qquad\qquad English Question & Question: Roger has 5 tennis balls. He buys 2 more cans of tennis balls. Each can has 3 tennis balls. How many tennis balls does he have now? \\
    \midrule    
    \translateen & Step-by-Step Answer: Roger started with 5 balls. 2 cans of 3 tennis balls each is 6 tennis balls. 5 + 6 = 11. The answer is 11. \\
    \bottomrule 
    \end{tabular}
    \caption{Example solution formats (\S\ref{sec:approaches}) for a German exemplar problem, where German-specific components are underlined and are changed to the corresponding translations for other investigated languages. For \direct, \nativecot and \encot, we provide the original German question as input to the model and expect an answer in the corresponding format; for \translateen, we input the translated question in English, and expect a step-by-step solution in English. 
    To obtain the desirable output format, we prepend few-shot examples in the corresponding format. 
    }
    \label{tab:prompt_example}
\end{table}

We provide an overview of standard prompting and chain-of-thought prompting, as well as their extensions to the multilingual setting, which we illustrate in Table~\ref{tab:prompt_example} and use in our experiments (\cref{sec:experiments}).

In standard prompting, given a prompt in the source language, the model is asked to predict the answer \citep{brown2020language,schick-schutze-2021-just}. 
This can be done in a zero-shot or few-shot setting by providing exemplars following the same template as additional input to the model. We refer to this setting as \textbf{direct answer prediction (\direct)} as the model directly predicts the answer to the problem. This setting measures the model's ability to solve problems without any intermediate reasoning steps.

Chain-of-thought \citep[\textsc{CoT};][]{wei2022chain} prompting helps improve many few-shot reasoning tasks, by augmenting few-shot examples with intermediate reasoning steps that should be predicted by the model. 
In the multilingual setting, we can apply CoT to \textbf{solve the problem in the native language (\nativecot)} by predicting the reasoning steps in the original language of the problem. This measures the model's ability to both understand and solve the problem in a specific language.

Alternatively, we can ask the model to \textbf{predict the chain of thought in English (\encot)}, regardless of the problem language. Such an approach may be useful as English is often used as the source language for cross-lingual transfer \citep{hu2020xtreme} and has been found effective when used as the prompt language \citep{Zhao2021,Winata2021,Lin2021}.

Finally, we can \textbf{translate the problem to English and solve it with English CoT (\translateen)}. 
In this setting, we use the Google Translate API to translate problems into English. 
This mirrors the translate-train setup \citep{hu2020xtreme,xue-etal-2021-mt5,ruder-etal-2021-xtreme}, the best-performing setting for fine-tuning multilingual models where the training data is translated to English. 

Beyond the prompting methods, there are different ways to provide few-shot examples in context for multilingual prompting:

\begin{itemize}[leftmargin=*]

\item \textbf{All native question exemplars (\nativeexemplar).} We use a few in-language questions together with their solutions as the few-shot prompt exemplars. 
This is the most natural setting when we have a few examples in each investigated language.

\item \textbf{All English question exemplars (\enexemplar).} When we are unable to access any existing questions or solution examples in some languages, an intuitive way is to use English questions and solutions as exemplars to perform zero-shot cross-lingual transfer. 
Note that it is unrealistic to combine this exemplar selection setting with \nativecot, since we assume no access to the native language for prompting. 

\item \textbf{Generic multilingual question exemplars (\multilingualexemplar).} Similar to \enexemplar, we assume access to questions and solutions in a few languages, and test if multilingual exemplars better elicit the multilingual reasoning ability of models.

\end{itemize}

\begin{figure}[!t]
\centering
\includegraphics[width=1.0\linewidth]{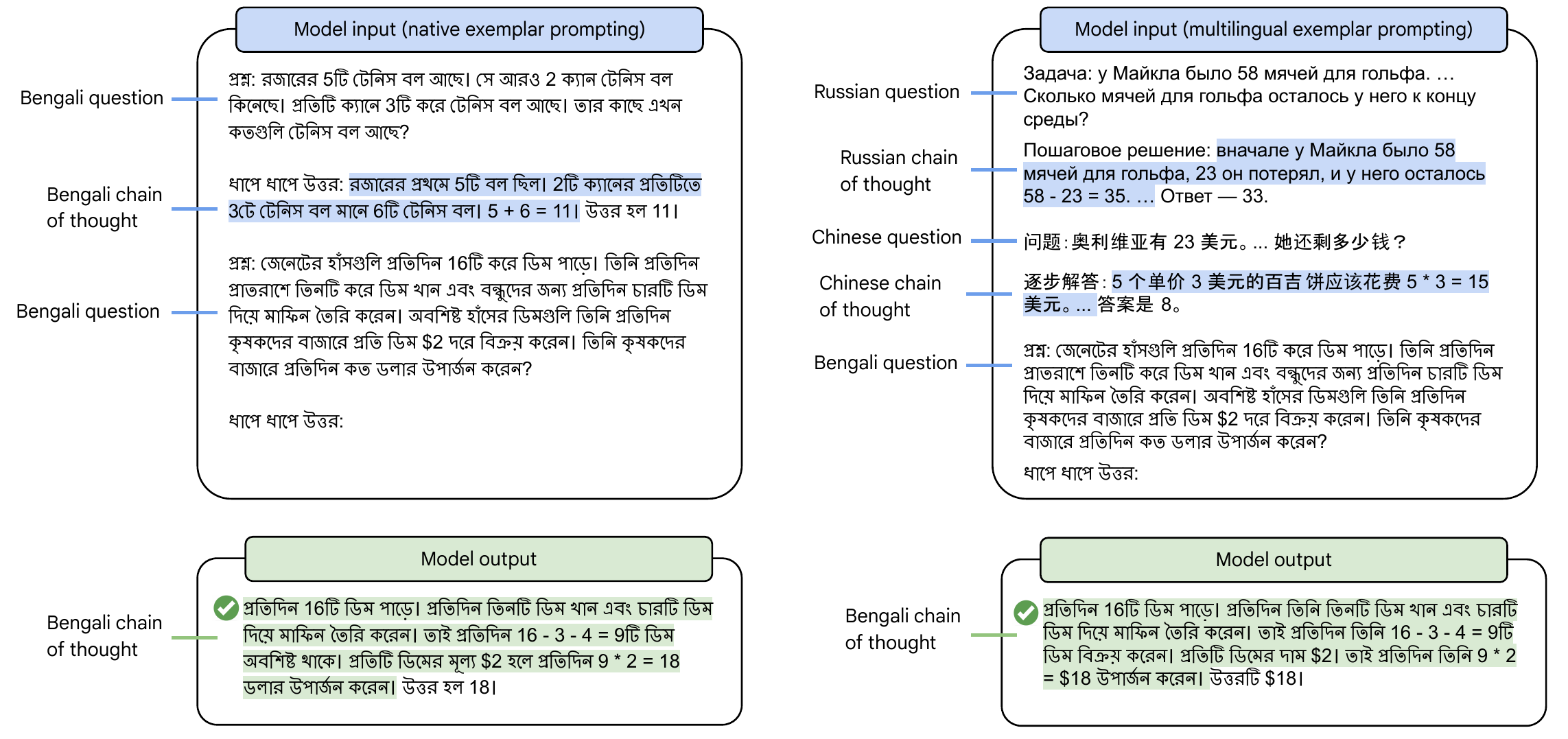}
\caption{
The chain-of-thought prompts and example model outputs in the \mgsm{} experiments.
The solutions are written in the same language as the questions of interest (\nativecot).
}
\label{fig:mgsm-prompts}
\end{figure}

For \translateen, as all exemplar questions and solutions are in English, we only experiment with the translated native question exemplars and English CoT.
We summarize the combinations of prompting and exemplar methods in \cref{tab:format_combination}, and present an illustration in \cref{fig:mgsm-prompts}.
Detailed prompting input for each investigated combination can be found in \cref{sec:detailed-mgsm-prompts}.

\begin{table}[t]
    \centering \small
    \begin{tabular}{lcccc}
        \toprule 
         & \direct & \nativecot &\encot & \translateen \\
        \midrule 
        \nativeexemplar & \checkmark & \checkmark & \checkmark & \checkmark \\ 
        \enexemplar & \checkmark & N/A & \checkmark & N/A \\
        \multilingualexemplar & \checkmark & \checkmark & \checkmark & N/A \\ 
        \bottomrule
    \end{tabular}
    \caption{Possible combinations between few-shot exemplar selection and solution strategies.}
    \label{tab:format_combination}
\end{table}

\section{Experiments on \mgsm} \label{sec:experiments}
\newcommand{\maintablecaption}{
Accuracy (\%) on \mgsm of different models and languages with exemplar questions in native languages (\nativeexemplar).
\textsc{hrl}: average performance across high-resource languages with larger than 0.1\% frequency in the training corpora; \textsc{url}: average performance across underrepresented languages.
We use 6 questions and solutions as the few-shot exemplar whenever possible: while the token number for 6-shot prompts in some languages may exceed the token number limit of GPT-3, we use the maximum possible number of exemplars instead for these cases. 
Detailed numbers of exemplars for each language in GPT-3 experiments can be found in \cref{sec:appendix:n-shots}. 
The best numbers in each column are in \textbf{boldface}. 
}

\begingroup
\setlength{\tabcolsep}{2.9pt}
\begin{table}[t]
    \centering
    \small
    \begin{tabular}{l rrr | r  | rrrrrr | rrrr}
    \toprule
    & \textsc{avg} & \textsc{hrl} & \textsc{url} & \multicolumn{1}{c|}{\textsc{en}} & \multicolumn{1}{c}{\textsc{de}} & \multicolumn{1}{c}{\textsc{fr}} & \multicolumn{1}{c}{\textsc{es}} & \multicolumn{1}{c}{\textsc{ru}} & \multicolumn{1}{c}{\textsc{zh}} & \multicolumn{1}{c|}{\textsc{ja}} & \multicolumn{1}{c}{\textsc{th}} & \multicolumn{1}{c}{\textsc{te}} & \multicolumn{1}{c}{\textsc{bn}} & \multicolumn{1}{c}{\textsc{sw}} \\
    \midrule
    Lang. Freq. (\PaLM, \%) & \multicolumn{1}{c}{--} & \multicolumn{1}{c}{--} & \multicolumn{1}{c|}{--} & 78.0 & 3.5 & 3.3 & 2.1 & .53 & .40 & .38 & .04 & .02 & .006 & .005 \\
    \midrule
    \multicolumn{10}{l}{\underline{GPT-3 (\openai{text}{davinci}{002})} \vspace{0.5mm}} \\
    \smallbullet \direct & 11.7 & 15.1 & 5.7 & 16.0 & 14.8 & 16.8 & 17.2 & 12.4 & 18.0 & 11.2 & 8.8 & 0.8 & 4.4 & 8.8 \\
    \smallbullet \nativecot & 26.4 & 34.7 & 7.2 & 53.6 & 36.0 & 37.6 & 40.4 & 28.4 & 40.0 & 26.0 & 10.8 & 0.4 & 6.4 & 11.2\\ 
    \smallbullet \encot & 31.6 & 39.4 & 13.9 & 53.6 & 44.0 & 46.0 & 44.8 & 28.4 & 40.8 & 32.4 & 19.6 & 5.6 & 9.6 & 20.8\\
    \smallbullet \translateen & 45.6 & 47.5 & 40.7 & 53.6 & 46.4 & 46.4 & 51.6 & 48.8 & 47.2 & 44.8 & 41.2 & 42.8 & 41.2 & 37.6 \\
    \midrule
    \underline{\PaLM-540B}  \vspace{0.5mm} \\
    \smallbullet \direct & 18.6 & 19.3 & 16.8 & 22.0 & 18.8 & 19.6 & 20.0 & 22.0 & 19.2 & 16.0 & 16.8 & 17.6 & 17.2 & 15.6 \\
    \smallbullet \nativecot & 48.1 & 47.9 & 44.9 & \bf 62.4 & 49.2 & 46.4 & 56.8 & 48.4 & 46.8 & 40.0 & 52.8 & 45.6 & 46.0 & 35.2 \\
    \smallbullet \encot & 51.3 & 52.3 & 46.8 & \bf 62.4 & 53.6 & 51.2 & 58.0 & 55.6 & 46.0 & 49.6 & 49.6 & 46.8 & 46.4 & 44.4 \\
    \smallbullet \translateen & \bf 55.0 & \bf 56.3 & \bf 51.2 & \bf 62.4 & \bf 57.2 & \bf 55.2 & \bf 60.0 & \bf 59.6 & \bf 55.6 & \bf 50.0 & \bf 50.8 & \bf 49.6 & \bf 53.2 & \bf 51.2 \\
    \bottomrule
    \end{tabular}
    \caption{\maintablecaption}
    \label{tab:main_table}
\end{table}
\endgroup

In this section, we evaluate the multilingual reasoning abilities of two representative state-of-the-art pretrained large language models---GPT-3 \citep{brown2020language} and \PaLM \citep{chowdhery2022palm}
---on our \mgsm benchmark in various prompting settings using exemplars in the source language (\nativeexemplar).\footnote{We focus on these two models due to their notable few-shot performance. In contrast, current multilingual models perform poorly in few-shot settings and are generally used for finetuning with more data \citep{Winata2021}.}
Throughout this paper, we generate outputs using greedy decoding (i.e., sampling with temperature $\tau=0$).

\subsection{Main Results}
We first compare the few-shot \nativeexemplar performance with different solution strategies (\cref{tab:main_table}).
In line with the English results reported by \citet{wei2022chain}, we find that intermediate reasoning steps (\nativecot and \encot) help both models achieve substantial reasoning performance gains across all languages, outperforming direct answer prediction with no explicit reasoning steps (\direct) by a significant margin. 
\PaLM shows exceptional multilingual reasoning ability: while it outperforms \InstructGPT on all languages with different settings, \PaLM-540B with intermediate reasoning steps (\nativecot and \encot) achieves results similar to \translateen on all languages, even on underrepresented languages such as Bengali (\textsc{bn}) and Swahili (\textsc{sw}), which cover less than 0.01\% of the training corpora. 

In addition, reasoning in English (\encot) consistently achieves competitive or better performance than reasoning in the native language of the question (\nativecot), suggesting that English intermediate steps can be considered as useful baseline in future work on multilingual reasoning. 

\subsection{Further Analysis}
\label{sec:mgsm-analysis}
\paragraph{Effect of language frequency in training corpora.}
We illustrate the main results of \nativecot, \encot and \translateen with respect to the language frequency in \PaLM training data (\cref{fig:freq_acc}).
Surprisingly, there is no strong correlation between the performance and the language frequency in the training corpora: the average accuracy among the four underrepresented languages was only 3\% lower than the that among the six high-resource languages (44.9\% vs 47.9\%).
Moreover, the performance of reasoning in Thai, Telugu, and Bengali is on par with reasoning in French, Japanese, and Chinese, despite having significantly much less data in the training corpora. 

In contrast to prior work that identifies language frequency as important for complex NLU tasks with relatively smaller models \citep{hu2020xtreme,Lauscher2020,ahuja-etal-2022-multi}, these results thus indicate that the reasoning ability of large language models may not be primarily dependent on their presence in training data and that language models are able to transfer their knowledge from high-resource to underrepresented languages to some extent. 

\begin{figure}[t]
\begin{minipage}{0.55\textwidth}
    \begin{centering}
    \begin{tikzpicture}
        \pgfplotsset{footnotesize,samples=10}
        \begin{groupplot}[
            group style = {group size = 2 by 1, horizontal sep = 20pt},
            width = 4.0cm, 
            height = 5cm]
            \nextgroupplot[
                align = center,
                title = {GPT-3},
                legend style={at={(-0.12,1.4)},anchor=south},
                xmin=0.5, xmax=5.5,
                ymin=-4, ymax=72,
                xtick={1, 2, 3, 4, 5},
                xticklabels={A, B, C, D$_1$, D$_2$},
                axis x line*=bottom,
                axis y line*=left,
                xlabel={Model Names},
                ylabel={\mgsm{} Accuracy (\%)},
                ytick={0, 10, 20, 30, 40, 50, 60, 70},
                grid style=dashed,
                x label style={at={(axis description cs:0.5,-0.15)},anchor=north},
                y label style={at={(axis description cs:-0.16,0.5)},anchor=south},
                xtick pos=bottom,
                ytick pos=left,
                ]
                \addplot[
                    color=black,
                    mark=*,
                    mark size=1pt,
                    line width=1pt,
                    ]
                    coordinates {
                    (1, 0.8)
                    (2, 2.8)
                    (3, 0.8)
                    (4, 15.2)
                    (5, 53.6)
                    };
                \addplot[
                    color=olive,
                    mark=*,
                    mark size=1pt,
                    line width=1pt,
                    ]
                    coordinates {
                    (1, 0.8)
                    (2, 1.6)
                    (3, 3.2)
                    (4, 12.8)
                    (5, 36)
                    };
                \addplot[
                    color=brown,
                    mark=*,
                    mark size=1pt,
                    line width=1pt,
                    ]
                    coordinates {
                    (1, 4.4)
                    (2, 1.2)
                    (3, 2.4)
                    (4, 12.4)
                    (5, 37.6)
                    };
                \addplot[
                    color=orange,
                    mark=*,
                    mark size=1pt,
                    line width=1pt,
                    ]
                    coordinates {
                    (1, 1.6)
                    (2, 1.2)
                    (3, 1.2)
                    (4, 10.8)
                    (5, 40.4)
                    };
                \addplot[
                    color=cyan,
                    mark=*,
                    mark size=1pt,
                    line width=1pt,
                    ]
                    coordinates {
                    (1, 2)
                    (2, 1.6)
                    (3, 0.8)
                    (4, 2.4)
                    (5, 10.4)
                    };
            \nextgroupplot[
                align = center,
                title = {\PaLM},
                legend style={at={(1,0.5)},anchor=west,draw=none},
                xmode=log,
                xmin=3, xmax=1200,
                ymin=-4, ymax=72,
                xtick={0.00001, 0.0001, 0.001, 0.01, 0.1, 1},
                axis x line*=bottom,
                axis y line*=left,
                xtick={8, 62, 540},
                xticklabels={8B, 62B, 540B},
                xlabel={Model Scale},
                ytick={0, 10, 20, 30, 40, 50, 60, 70},
                grid style=dashed,
                x label style={at={(axis description cs:0.5,-0.15)},anchor=north},
                y label style={at={(axis description cs:-0.16,0.5)},anchor=south},
                xtick pos=bottom,
                ytick pos=left,
                ]
                \addplot[
                    color=black,
                    mark=*,
                    mark size=1pt,
                    line width=1pt,
                    ]
                    coordinates {
                    (8, 6.4)
                    (62, 30.4)
                    (540, 62.4)
                    };
                    \addlegendentry{en}
                \addplot[
                    color=olive,
                    mark=*,
                    mark size=1pt,
                    line width=1pt,
                    ]
                    coordinates {
                    (8, 6.8)
                    (62, 24)
                    (540, 49.2)
                    };
                    \addlegendentry{de}
                \addplot[
                    color=brown,
                    mark=*,
                    mark size=1pt,
                    line width=1pt,
                    ]
                    coordinates {
                    (8, 4.4)
                    (62, 24.0)
                    (540, 46.4)
                    };
                    \addlegendentry{fr}
                \addplot[
                    color=orange,
                    mark=*,
                    mark size=1pt,
                    line width=1pt,
                    ]
                    coordinates {
                    (8, 2.4)
                    (62, 26.0)
                    (540, 56.8)
                    };
                    \addlegendentry{es}
                \addplot[
                    color=red,
                    mark=*,
                    mark size=1pt,
                    line width=1pt,
                    ]
                    coordinates {
                    (8, 2.8)
                    (62, 22.8)
                    (540, 48.4)
                    };
                    \addlegendentry{ru}
                \addplot[
                    color=purple,
                    mark=*,
                    mark size=1pt,
                    line width=1pt,
                    ]
                    coordinates {
                    (8, 4.0)
                    (62, 24.8)
                    (540, 46.8)
                    };
                    \addlegendentry{zh}
                \addplot[
                    color=magenta,
                    mark=*,
                    mark size=1pt,
                    line width=1pt,
                    ]
                    coordinates {
                    (8, 4.4)
                    (62, 14.8)
                    (540, 40.0)
                    };
                    \addlegendentry{ja}
                \addplot[
                    color=violet,
                    mark=*,
                    mark size=1pt,
                    line width=1pt,
                    ]
                    coordinates {
                    (8, 3.2)
                    (62, 18.0)
                    (540, 52.8)
                    };
                    \addlegendentry{th}
                \addplot[
                    color=blue,
                    mark=*,
                    mark size=1pt,
                    line width=1pt,
                    ]
                    coordinates {
                    (8, 3.6)
                    (62, 11.6)
                    (540, 45.6)
                    };
                    \addlegendentry{te}
                \addplot[
                    color=teal,
                    mark=*,
                    mark size=1pt,
                    line width=1pt,
                    ]
                    coordinates {
                    (8, 3.2)
                    (62, 13.6)
                    (540, 46.0)
                    };
                    \addlegendentry{bn}
                \addplot[
                    color=cyan,
                    mark=*,
                    mark size=1pt,
                    line width=1pt,
                    ]
                    coordinates {
                    (8, 2.4)
                    (62, 9.6)
                    (540, 35.2)
                    };
                    \addlegendentry{sw}
        \end{groupplot}
    \end{tikzpicture}
    \caption{
    \mgsm accuracy with different model scales.
    The letters A, B, C, D$_1$, and D$_2$ denote \openai{text}{ada}{001}, \openai{text}{babbage}{001}, \openai{text}{curie}{001}, \openai{text}{davinci}{001}, and \openai{text}{davinci}{002} in the GPT-3 \citep{brown2020language,ouyang2022training} family, respectively.
    While the number of parameters in each GPT-3 model is not publicly available, we order them alphabetically. 
    Detailed numbers can be found in \cref{tab:appendix_table}.
    } 
    \label{fig:scaling_laws}
    \end{centering}
\end{minipage}
\hfill
\begin{minipage}{0.42\textwidth}
    \begin{centering}
    \begin{tikzpicture}
        \pgfplotsset{footnotesize,samples=10}
        \begin{groupplot}[
            group style = {group size = 1 by 1, horizontal sep = 20pt},
            width = 5cm, 
            height = 5cm]
            \nextgroupplot[
                align = center,
                legend style={at={(1,0.5)},anchor=west,draw=none},
                xmin=0.5, xmax=7,
                ymin=-4, ymax=72,
                xtick={0.00001, 0.0001, 0.001, 0.01, 0.1, 1},
                axis x line*=bottom,
                axis y line*=left,
                xtick={1, 2, 4, 6},
                xlabel={\# few-shot exemplars},
                ylabel={\mgsm{} Accuracy (\%)},
                ytick={0, 10, 20, 30, 40, 50, 60, 70},
                grid style=dashed,
                x label style={at={(axis description cs:0.5,-0.15)},anchor=north},
                y label style={at={(axis description cs:-0.16,0.5)},anchor=south},
                xtick pos=bottom,
                ytick pos=left,
                ]
                \addplot[
                    color=black,
                    mark=*,
                    mark size=1pt,
                    line width=1pt,
                    ]
                    coordinates {
                    (1, 50.8)
                    (2, 57.2)
                    (4, 58.8)
                    (6, 62.4)
                    };
                    \addlegendentry{en}
                \addplot[
                    color=olive,
                    mark=*,
                    mark size=1pt,
                    line width=1pt,
                    ]
                    coordinates {
                    (1, 42.8)
                    (2, 47.2)
                    (4, 44.8)
                    (6, 49.2)
                    };
                    \addlegendentry{de}
                \addplot[
                    color=brown,
                    mark=*,
                    mark size=1pt,
                    line width=1pt,
                    ]
                    coordinates {
                    (1, 44.9)
                    (2, 43.2)
                    (4, 49.6)
                    (6, 46.4)
                    };
                    \addlegendentry{fr}
                \addplot[
                    color=orange,
                    mark=*,
                    mark size=1pt,
                    line width=1pt,
                    ]
                    coordinates {
                    (1, 44.8)
                    (2, 50.4)
                    (4, 47.6)
                    (6, 56.8)
                    };
                    \addlegendentry{es}
                \addplot[
                    color=red,
                    mark=*,
                    mark size=1pt,
                    line width=1pt,
                    ]
                    coordinates {
                    (1, 41.2)
                    (2, 44.4)
                    (4, 46.4)
                    (6, 48.4)
                    };
                    \addlegendentry{ru}
                \addplot[
                    color=purple,
                    mark=*,
                    mark size=1pt,
                    line width=1pt,
                    ]
                    coordinates {
                    (1, 34.8)
                    (2, 44.4)
                    (4, 46.4)
                    (6, 46.8)
                    };
                    \addlegendentry{zh}
                \addplot[
                    color=magenta,
                    mark=*,
                    mark size=1pt,
                    line width=1pt,
                    ]
                    coordinates {
                    (1, 29.6)
                    (2, 34.4)
                    (4, 38.4)
                    (6, 40.0)
                    };
                    \addlegendentry{ja}
                \addplot[
                    color=violet,
                    mark=*,
                    mark size=1pt,
                    line width=1pt,
                    ]
                    coordinates {
                    (1, 40.0)
                    (2, 47.2)
                    (4, 46.4)
                    (6, 52.8)
                    };
                    \addlegendentry{th}
                \addplot[
                    color=blue,
                    mark=*,
                    mark size=1pt,
                    line width=1pt,
                    ]
                    coordinates {
                    (1, 38.0)
                    (2, 38.0)
                    (4, 41.2)
                    (6, 45.6)
                    };
                    \addlegendentry{te}
                \addplot[
                    color=teal,
                    mark=*,
                    mark size=1pt,
                    line width=1pt,
                    ]
                    coordinates {
                    (1, 34.0)
                    (2, 40.8)
                    (4, 44.4)
                    (6, 46.0)
                    };
                    \addlegendentry{bn}
                \addplot[
                    color=cyan,
                    mark=*,
                    mark size=1pt,
                    line width=1pt,
                    ]
                    coordinates {
                    (1, 27.2)
                    (2, 33.2)
                    (4, 32.0)
                    (6, 35.2)
                    };
                    \addlegendentry{sw}
        \end{groupplot}
    \end{tikzpicture}
    \caption{
    \mgsm accuracy of \PaLM-540B with different numbers of few-shot exemplars.
    Detailed numbers can be found in \cref{tab:appendix_table}.
    } 
    \label{fig:num_shots}
    \end{centering}
\end{minipage}
\end{figure}

\paragraph{Effect of model scale.}
We analyze the effect of model scale (i.e., number of model parameters and computational resources used for training) on their multilingual arithmetic reasoning abilities (\cref{fig:scaling_laws}). 
As the models scale up, the performance generally improves for both GPT-3 and \PaLM model series on all languages.
Neither model achieves a substantial solve rate until a certain scale (\openai{text}{davinci}{001} for GPT-3 and \PaLM-62B for \PaLM), hence multilingual reasoning can be considered an \textit{emergent ability} of large language models \citep{wei2022emergent}. 
It is worth noting that the amount of training data per language is constant across language model scales for \PaLM---the fact that scale facilitates reasoning implies that further scaling may continue to improve the multilingual reasoning ability of large language models.

\paragraph{Effect of exemplar amount.}
We analyze how the multilingual reasoning performance of \PaLM-540B, the overall best-performing model, is affected by the number of few-shot exemplars (\cref{fig:num_shots}). 
Although not all trends are strictly increasing with the number of exemplars, \PaLM-540B benefits from having more examples in general for all languages.

\paragraph{Effect of exemplar type choice.}

\begingroup
\setlength{\tabcolsep}{2.9pt}
\begin{table}[t]
    \centering
    \small
    \begin{tabular}{l rrr | r  | rrrrrr | rrrr}
    \toprule
     & \textsc{avg} & \textsc{hrl} & \textsc{url} & \multicolumn{1}{c|}{\textsc{en}} & \multicolumn{1}{c}{\textsc{de}} & \multicolumn{1}{c}{\textsc{fr}} & \multicolumn{1}{c}{\textsc{es}} & \multicolumn{1}{c}{\textsc{ru}} & \multicolumn{1}{c}{\textsc{zh}} & \multicolumn{1}{c|}{\textsc{ja}} & \multicolumn{1}{c}{\textsc{th}} & \multicolumn{1}{c}{\textsc{te}} & \multicolumn{1}{c}{\textsc{bn}} & \multicolumn{1}{c}{\textsc{sw}} \\
    \midrule
    \multicolumn{10}{l}{\nativeexemplar} \\
    \midrule     
    \nativecot & 48.1 & 47.9 & 44.9 & \bf 62.4 & 49.2 & 46.4 & 56.8 & 48.4 & 46.8 & 40.0 & 52.8 & 45.6 & 46.0 & 35.2 \\
    \encot &\bf  51.3 & \bf 52.3 & \bf 46.8 & \bf 62.4 & \bf 53.6 & \bf 51.2 & \bf 58.0 & \bf 55.6 & 46.0 & \bf49.6 & \bf49.6 & \bf46.8 & 46.4 & \bf44.4 \\
    \midrule \midrule 
    \multicolumn{10}{l}{\multilingualexemplar} \\
    \midrule 
    \nativecot & 29.8 & 31.8 & 26.3 & 52.0 & 41.6 & 7.2 & 10.4 & 36.0 & 42.8 & 32.8 & 18.0 & 33.6 & 26.8 & 26.8\\
    \encot & 48.7 & 50.0 & 46.3 & 57.6 & 53.2 & 43.2 & 53.2 & 48.0 & \bf 51.2 & 43.6 & 46.8 & 46.4 & \bf 48.4 & 43.6\\
    \midrule \midrule
    \multicolumn{10}{l}{\enexemplar} \\
    \midrule
    \encot & 34.7 & 39.4 & 26.6 & \bf 62.4 & 46.0 & 37.2 & 50.4 & 23.6 & 29.2 & 26.8 & 17.2 & 30.0 & 34.4 & 24.8\\
    \bottomrule
    \end{tabular}
    \caption{Performance on \mgsm with different prompt exemplar type choices: the first section is copied correspondingly from \cref{tab:main_table}. The best numbers in each column are in \textbf{boldface}. }
    \label{tab:exemplar-choice}
\end{table}
\endgroup
We compare the multilingual reasoning performance of \PaLM-540B across languages with different exemplar choices (\cref{tab:exemplar-choice}). 
For the \multilingualexemplar setting, we concatenate one example from each of the most frequent languages (English, German, French, Spanish, Russian, and Chinese) as the generic prompt for all languages. 
While the best choice is almost always to use \nativeexemplar and \encot, \multilingualexemplar with \encot achieves competitive performance across the board, suggesting an effective approach when we do not have access to any existing example in some languages. 

Most notably, with \encot, \multilingualexemplar significantly outperforms \enexemplar on all non-English languages, including those not covered by the few-shot examples, suggesting that a multilingual few-shot prompt helps elicit the multilingual reasoning abilities of models more effectively than a monolingual (English) one.

\section{Extension to Other Multilingual Reasoning Benchmarks}
\label{sec:expr-others}
To better understand the multilingual reasoning abilities of large pretrained language models, we extend our experiments to two additional multilingual reasoning benchmarks, \xcopa \citep{ponti-etal-2020-xcopa} and \xlwic \citep{raganato-etal-2020-xl}. 
Throughout this section, we evaluate the Codex \citep[\openai{code}{davinci}{002};][]{chen2021evaluating}\footnote{For both investigated tasks, we find that \openai{code}{davinci}{002} generally produces competitive or better results than \openai{text}{davinci}{002} on a small set of samples. In consideration of budget, we choose to use \openai{code}{davinci}{002} because it supports free access at the time of our experiment. } and \PaLM-540B models.

\subsection{\xcopa}
\label{sec:results-xcopa}

\begin{table}[!t]
\small 
\centering
\setlength{\tabcolsep}{3.8pt}
\begin{tabular}{l|c|ccccccccccc}
\toprule
\textbf{\textsc{Model}} & \textbf{\textsc{avg}} & \textsc{et} & \textsc{ht} & \textsc{id} & \textsc{it} & \textsc{qu} & \textsc{sw} & \textsc{ta} & \textsc{th} & \textsc{tr} & \textsc{vi} & \textsc{zh} \\
\midrule
\textsc{Human}  & 97.6 & 98.2 & 96.4 & 100 & 97 & 94.8 & 99 & 98.6 & 98.2 & 96.4 & 98.4 & 96.6 \\
\midrule
MAD-X Base  & 61.0 & 61.3 & 53.7 & 65.8 & 63.0 & 52.5 & 56.3 & 61.9 & 61.8 & 60.3 & 66.1 & 67.6 \\
XLM-R Large & 68.7 & 71.4 & (50) & 79.8 & 72.6 & (50) & 59.2 & 73 & 72.8 & 74.4 & 73.8 & 78.6 \\ 
mT5-XXL & 74.9 & 77.5 & 72.1 &  81.1 &  75.9 &  54.5 &  74.1 & 75.9 & 78.3 & 78.1 & 76.9 & 79.5 \\
RoBERTa Large (TT)   & 76.1 & 81.0 & 73.8 & 82.2 & 77.8 & (50) & 74.2 & 79.6 & 71.4 & 79.6 & 81.0 & 86.0 \\
\cdashlinelr{1-13}
\multicolumn{13}{l}{\underline{Codex (\openai{code}{davinci}{002})} \vspace{0.5mm}} \\
\smallbullet \textsc{Direct} & 73.3 & 73.8 & 55.6 & 88.8 & 95.4 & 51.2 & 56.0 & 54.6 & 70.2 & 88.6 & 80.4 & 91.4 \\
\smallbullet \textsc{En-CoT} & 80.7 & 88.8 & 79.6 & 91.4 & 96.6 & 52.2 & 67.4 & 55.8 & 84.2 & 91.2 & 86.6 & 93.4 \\
\cdashlinelr{1-13}
\multicolumn{13}{l}{\underline{\PaLM-540B} \vspace{0.5mm}} \\
\smallbullet \textsc{Direct} & 83.7 & 77.4 & 78.0 & 92.6 & 96.0 & 61.0 & 69.4 & 85.4 & 87.2 & 92.8 & 89.8 & 91.6 \\
\smallbullet \textsc{En-CoT} & \textbf{89.9} & \textbf{91.0} & \textbf{89.6} & \textbf{94.0} & \textbf{97.4} & \textbf{66.8} & \textbf{85.4} & \textbf{90.8} & \textbf{90.2} & \textbf{94.6} & \textbf{94.6} & \textbf{94.8} \\
\bottomrule
\end{tabular}
\caption{Accuracy on the \xcopa languages compared to previous work. Human evaluation (\textsc{Human}) on \xcopa was performed by \citet{ponti-etal-2020-xcopa}. The MAD-X Base, XLM-R Large, and RoBERTa Large (\emph{translate test}) results are from \citet{ponti-etal-2020-xcopa}, whereas the mT5 results are from \citep{ruder-etal-2021-xtreme}. Applying multilingual CoT-prompting to \PaLM-540B has enabled us to achieve a new state-of-the-art performance on \xcopa. The best model result in each column is in \textbf{boldface}.}
\label{tab:xcopa_results}
\end{table}

\xcopa is a multilingual evaluation dataset designed to assess the causal commonsense reasoning capabilities of language models across multiple languages.\footnote{\url{https://github.com/cambridgeltl/xcopa}} It is an extension and re-annotation of the English COPA dataset \citep{gordon-etal-2012-semeval} where the validation and test set examples are carefully translated to and annotated in 11 typologically diverse languages. These languages are Estonian (\textsc{et}), Indonesian (\textsc{id}), Italian (\textsc{it}), Cusco-Collao Quechua (\textsc{qu}), Swahili (\textsc{sw}), Tamil (\textsc{ta}), Thai (\textsc{th}), Turkish (\textsc{tr}), Vietnamese (\textsc{vo}), and Mandarin Chinese (\textsc{zh}). The task objective is to determine the causal relationship between the premise and two options based on a question (which is either ``What was the \emph{cause}?'' or ``What happened as a \emph{result}?''). A successful model is, therefore, expected to not only perform commonsense reasoning but also generalize its reasoning capabilities to new languages. For each target language, \xcopa contains $100$ annotated examples in the validation set and $500$ examples in the test set. In our experiments, we focus on the examples in the test sets and use the ones in the validation set as few-shot exemplars whenever needed.

We test the Codex and \PaLM models under both \direct and \encot. In both settings, we include the same set of examples, randomly selected from the validation sets of \textsc{tr}, \textsc{zh}, \textsc{ta}, and \textsc{qu}, but for \encot, we additionally write brief rationales (in English) before the final answers ourselves.

\paragraph{Results.} 

Table~\ref{tab:xcopa_results} presents our main results, along with per-language breakdowns for each \xcopa language. The previous state-of-the-art performance was around $76\%$, obtained by RoBERTa Large in the translate-test setting where the English RoBERTa Large model was first trained on the English COPA \citep{gordon-etal-2012-semeval} and English SIQa 
\citep{sap-etal-2019-social} datasets and then applied to the XCOPA test data, which was translated to English \citep{ponti-etal-2020-xcopa}. With only four multilingual chain-of-thought examples (\encot), \PaLM-540B outperforms RoBERTa Large by a significant margin ($14\%$), thereby setting a new high bar on \xcopa. While Codex performs better than RoBERTa Large, it still falls $9\%$ behind \PaLM-540B. We also highlight that \PaLM-540B performs noticeably better than all the other models on under-represented languages such as \textsc{et}, \textsc{ht}, and \textsc{sw}; this result suggests that \PaLM-540B might have some internal knowledge about these languages.

\subsection{\xlwic}
\label{sec:results-xlwic}

\begin{table}[t]
\setlength{\tabcolsep}{2.9pt}
\small 
\centering
\begin{tabular}{l|c|rrrrrrrrrrrr}
\toprule
\textbf{Model} & \textbf{\textsc{avg}} & \multicolumn{1}{c}{\textsc{bg}} & \multicolumn{1}{c}{\textsc{da}}& \multicolumn{1}{c}{\textsc{de}}& \multicolumn{1}{c}{\textsc{et}}& \multicolumn{1}{c}{\textsc{fa}}& \multicolumn{1}{c}{\textsc{fr}}& \multicolumn{1}{c}{\textsc{hr}}& \multicolumn{1}{c}{\textsc{it}}& \multicolumn{1}{c}{\textsc{ja}}& \multicolumn{1}{c}{\textsc{ko}}& \multicolumn{1}{c}{\textsc{nl}}& \multicolumn{1}{c}{\textsc{zh}} \\
\midrule
\textsc{Human} & \multicolumn{1}{c|}{--} & 87.0 & \multicolumn{1}{c}{--} & 74.0 & \multicolumn{1}{c}{--} &  97.0  & \multicolumn{1}{c}{--} & \multicolumn{1}{c}{--} & 78.0 & 75.0 & 76.0 & \multicolumn{1}{c}{--} & 85.0 \\
\midrule \midrule 
XLM-R Large & \bf 68.9 & \bf 66.5 & \bf 71.1 & 65.8 & \bf 68.7 & \bf 75.3 & 62.5 & \bf 72.3 & \bf 64.9 & 63.8 & 69.6 & \bf 72.8 & \bf 73.2 \\
\cdashlinelr{1-14}
\multicolumn{13}{l}{\underline{Codex (\openai{code}{davinci}{002})} \vspace{0.5mm}} \\
\direct & 60.8 &  59.2 & 59.6 & 68.2 & 59.0 & 58.0 & 58.6 & 65.7 & 55.4 & 56.0 & 62.0 & 64.8 & 63.0  \\
\encot & 61.4 &  60.2 & 66.6 & 70.6 & 60.3 & 63.6 & \bf 64.6 & 61.0 & 54.2 & 52.2 & 56.6 & 62.8 & 64.0 \\ 
\cdashlinelr{1-14}
\multicolumn{13}{l}{\underline{\PaLM-540B} \vspace{0.5mm}} \\
\direct & 66.7 & 62.6 & 67.4 & \bf 72.6 & 62.3 & 75.0 & \bf 64.6 & 65.0 & 59.4 & \bf 64.0 & \bf 70.2 & 72.0 & 64.8 \\ 
\encot & 63.2 & 63.4 & 64.6 & 68.6 & 61.5 & 67.2 & \bf 64.6 & 55.9 & 57.4 & 55.6 & 66.4 & 69.4 & 64.0 \\
\bottomrule
\end{tabular}
\caption{Accuracy on the \xlwic languages with \multilingualexemplar. XLM-R Large denotes the previous state-of-the-art results trained with 5.4K English examples \citep{raganato-etal-2020-xl}. The best model result in each column is in \textbf{boldface}. }
\label{tab:xlwic-results}
\end{table}

\xlwic is a multilingual word in-context semantic judgment benchmark covering thirteen languages:\footnote{\url{https://pilehvar.github.io/xlwic/}} Bulgarian (\textsc{bg}), Danish (\textsc{da}), German (\textsc{de}), Estonian (\textsc{et}), Persian (\textsc{fa}), French (\textsc{fr}), Croatian (\textsc{hr}), Italian (\textsc{it}), Japanese (\textsc{ja}), Korean (\textsc{ko}), Dutch (\textsc{nl}) and Chinese (\textsc{zh}). 
Given two sentences in the same language and a word of interest which appears in both sentences, the model is asked whether the word is of the same sense in the sentences. 
In order to arrive at the correct answer, a model needs to be aware of the concept of word sense, and to infer the sense of a word based on its context. Despite its simplicity, this task is extremely challenging; \PaLM-540B only achieves a score of 64.6 on WiC 
\citep{pilehvar-camacho-collados-2019-wic}, the English version of the task.

\paragraph{Results.} We evaluate the cross-lingual word-in-context sense judgment performance of models (\cref{tab:xlwic-results}). 
With the supervision from only four examples, \PaLM-540B achieves competitive or better results that the state-of-the-art model (XLM-R Large) on 6 (German, Persian, French, Japanese, Korean and Dutch) of the 12 investigated languages. 
However, we do not observe an improvement over direct answer prediction when using chain-of-thought prompting on this task.\footnote{One potential reason is that our prompts are not necessarily optimal \citep{wang2022rationale} and may benefit from a broader investigation of other prompt formats. 
On the other hand, rationales for this task are fairly straight-forward and example-specific. It is thus unclear whether the WiC task requires true reasoning that benefits from the depiction of intermediate reasoning steps. We leave further investigation for future work. }

\section{Related Work}
\paragraph{Prompting.}
Existing work \interalia{radford2019language,brown2020language,schick-schutze-2021-just} has shown that prompting pre-trained large language models can lead to strong performance on various tasks such as text classification \citep{shin2020autoprompt,gao-etal-2021-making}, question answering \citep{khashabi-etal-2020-unifiedqa}, and program synthesis \citep{austin2021program,nye2021show,shi2022natural}: taking a few examples of the task in a certain pattern as the prompting input, models are often able to generate accurate output following the pattern. 
\citet{wei2022chain} have shown that chain-of-thought prompting significantly improves the reasoning performance of language models, by adding explicit reasoning steps before the final answer. 
\citet{ahn2022can} apply chain-of-thought prompting in robotics scenarios, including a multilingual setting. 
In this work, we systematically analyze multilingual few-shot chain-of-thought prompting on complicated reasoning benchmarks. 

\paragraph{Multilingual pre-trained language models.}
Through masked language modeling \citep{devlin-etal-2019-bert,conneau-etal-2020-unsupervised}, auto-regressive language modeling \citep{brown2020language,ouyang2022training} or encoder-decoder training \citep{liu2020multilingual,chen2021evaluating,xue-etal-2021-mt5}, pre-trained Transformer-based large language models have shown impressive performance on multiple NLP tasks across languages. Previous work \citep{Zhao2021,Winata2021,Lin2021} investigated prompting in the multilingual setting and found that using English prompts with non-English examples led to strong few-shot performance. Evaluation of multilingual models has mostly focused on general information extraction tasks such as question answering \citep{clark-etal-2020-tydi,hu2020xtreme,Kassner2021,ruder2021multi} as well as specific types of reasoning such as commonsense reasoning \citep{ponti-etal-2020-xcopa,lin-etal-2021-common} and temporal reasoning \citep{ruder-etal-2021-xtreme}. To the best of our knowledge, this is the first study to evaluate the multilingual multi-step reasoning abilities of large language models.

\paragraph{Cross-lingual transfer and generalization.} 
Previous work has demonstrated that pre-trained multilingual models significantly help cross-lingual transfer on a wide range of NLP tasks such as cross-lingual named entity recognition \citep{pires-etal-2019-multilingual,mulcaire-etal-2019-polyglot}, zero-shot cross-lingual dependency parsing \citep{schuster-etal-2019-cross,shi-etal-2022-substructure}, and bilingual lexicon induction \citep{shi-etal-2021-bilingual}. 
In this work, we demonstrate strong cross-lingual generalization of \PaLM (\S\ref{sec:mgsm-analysis}, \S\ref{sec:expr-others}) and Codex (\S\ref{sec:expr-others}), on three tasks that require complicated reasoning.

\paragraph{Multilingual benchmarks.} 
To test the multilingual NLP performance of existing models, there has been work introducing benchmarks on various multilingual tasks, including cross-lingual question answering \citep{liu-etal-2019-xqa,clark-etal-2020-tydi}, natural language inference \citep{conneau-etal-2018-xnli} and bilingual lexicon induction \citep{lample2018word}, as well as collections across tasks \citep{hu2020xtreme,ruder-etal-2021-xtreme}. 
The tasks in these multilingual benchmarks, to the best of our knowledge, require relatively simple reasoning processes. 
In this paper, we present \mgsm, a multilingual arithmetic reasoning benchmark, which can be used to test multilingual multi-step reasoning abilities of models.

\section{Conclusion}
In this paper, we introduce \mgsm, the first multilingual benchmark to evaluate arithmetic reasoning abilities of language models. MGSM is an extension of the GSM8K dataset \citep{cobbe2021training} and contains 250 examples written in \emph{ten} typologically diverse languages. We also present a comprehensive analysis of the multilingual reasoning abilities of large language models such as GPT-3 and \PaLM on multiple multilingual benchmarks, including our own MGSM dataset. We find that large-scale language models appear to perform complex multi-step reasoning across multiple languages, including those underrepresented languages which are covered by less than $0.01\%$ of training corpora. Finally, we demonstrate that multilingual chain-of-thought prompting is an empirically effective approach to multilingual commonsense reasoning, outperforming the previous best model on the challenging XCOPA dataset by 13\% on average.

\clearpage
\bibliography{main}
\bibliographystyle{iclr2023_conference}

\clearpage
\appendix
\section{Details of \mgsm Experiments}
In this section, we present details of our experiments on \mgsm, including the number of exemplars used for GPT-3 (\S\ref{sec:appendix:n-shots}) and the detailed prompts in each setting summarized in \cref{tab:format_combination} (\S\ref{sec:detailed-mgsm-prompts}).
\subsection{Number of Exemplars for Each Language }
\label{sec:appendix:n-shots}
Given the unbalanced representation of languages in the training corpora, the byte-pair encoding \citep[BPE;][]{gage1994new} algorithm tokenizes sentences in underrepresented languages, especially those in a different alphabet from English, into more tokens. 
Given that the GPT-3 API supports a maximum number of 2048 tokens as its input, it does not support 6-shot prompting in some languages, including Russian, Chinese, Japanese, Thai, Telugu and Bengali; therefore, we use the maximum possible number of exemplars (\cref{tab:n_exemplar_per_language}) instead for GPT-3, while using 6-shot for all languages in \PaLM experiments.

\begingroup
\setlength{\tabcolsep}{2.9pt}
\begin{table}[t]
    \centering
    \small
    \begin{tabular}{l | rrrrrrrrrrr}
    \toprule
     & \multicolumn{1}{c}{en} & \multicolumn{1}{c}{de} & \multicolumn{1}{c}{fr} & \multicolumn{1}{c}{es} & \multicolumn{1}{c}{ru} & \multicolumn{1}{c}{zh} & \multicolumn{1}{c}{ja} & \multicolumn{1}{c}{th} & \multicolumn{1}{c}{te} & \multicolumn{1}{c}{bn} & \multicolumn{1}{c}{sw} \\
    \midrule
    \# Exemplars & 6 & 6 & 6 & 6 & 1 & 5 & 4 & 1 & 1 & 1 & 6 \\
    \bottomrule
    \end{tabular}
    \caption{Number of few-shot exemplars for GPT-3 experiments in \cref{tab:main_table}.}
    \label{tab:n_exemplar_per_language}
\end{table}
\endgroup

\subsection{\mgsm Prompts in Each Setting}
\label{sec:detailed-mgsm-prompts}
\begin{figure}
    \centering
    \includegraphics[width=1.\textwidth]{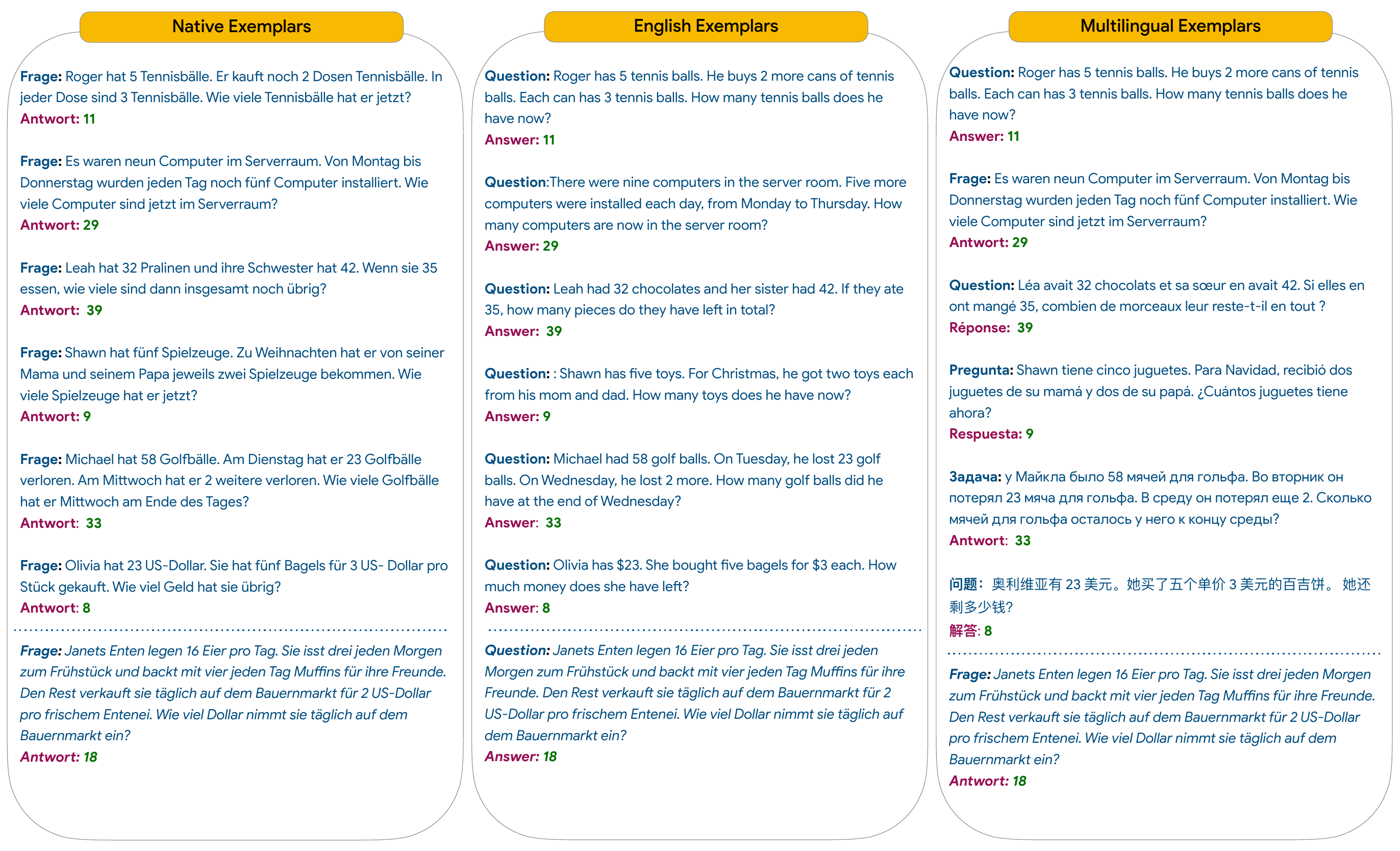}
    \caption{Prompt template in the direct answer prediction setting (\direct), solving a problem in German. Above dotted lines: few-shot exemplars; below dotted lines: the question of interest and the expected answer. The dotted lines are not included in our experiments.}
    \label{fig:direct-detailed-mgsm-prompt}
\end{figure}

\begin{figure}
    \centering
    \includegraphics[width=1.0\textwidth]{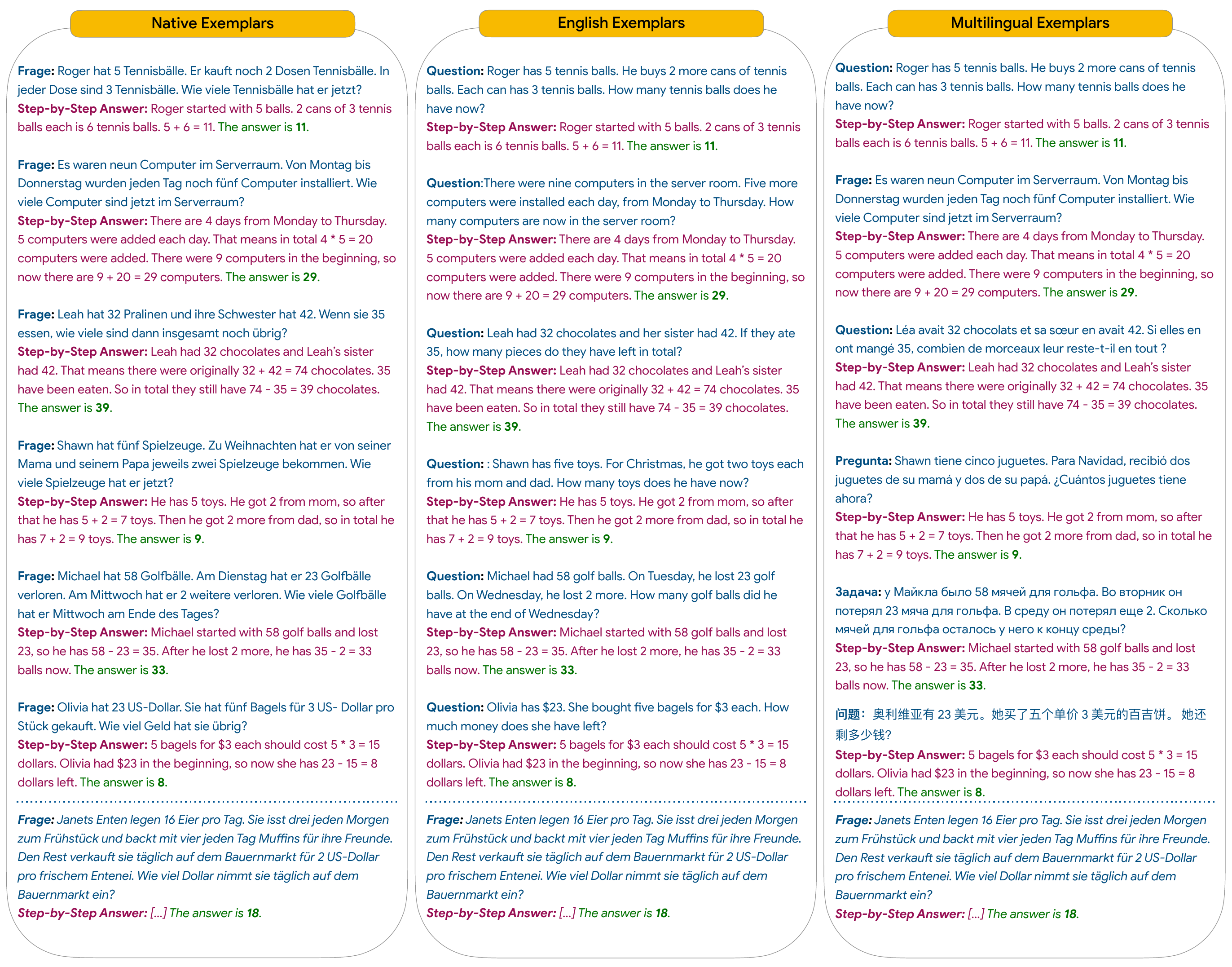}
    \caption{Prompt template in the English CoT setting (\encot), solving a problem in German. Above dotted lines: few-shot exemplars; below dotted lines: the question of interest and the expected answer. The dotted lines are not included in our experiments. }
    \label{fig:en-cot-detailed-mgsm}
\end{figure}

\begin{figure}
    \centering
    \includegraphics[width=1.0\textwidth]{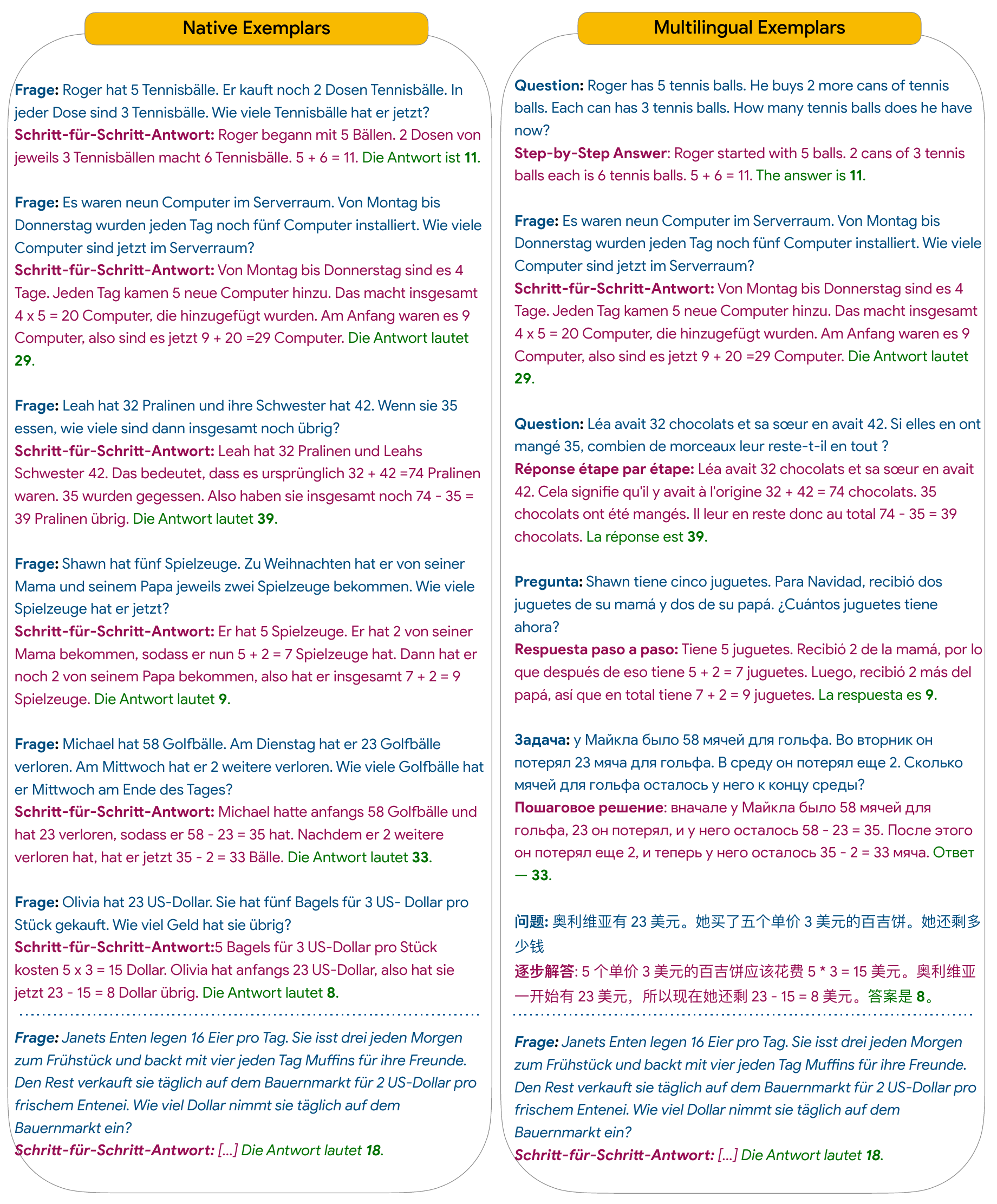}
    \caption{Prompt template with CoT in the question language (\nativecot), solving a problem in German. Above dotted lines: few-shot exemplars; below dotted lines: the question of interest and the expected answer. The dotted lines are not included in our experiments.}
    \label{fig:ml-cot-detailed-mgsm}
\end{figure}

We present the prompts used in our \mgsm experiments in \cref{fig:direct-detailed-mgsm-prompt,fig:en-cot-detailed-mgsm,fig:ml-cot-detailed-mgsm}, where the \translateen experiments can be viewed as a English one with \encot and \enexemplar.  

\section{Detailed MGSM Performance}
\label{sec:detailed_mgsm_analysis_numbers}
\begingroup
\setlength{\tabcolsep}{2.8pt}
\begin{table*}[t]
    \centering
    \small
    \begin{tabular}{l rrr | r  | rrrrrr | rrrr}
    \toprule
     & \textsc{avg} & \textsc{hrl} & \textsc{lrl} & \textsc{en} & \textsc{de} & \textsc{fr} & \textsc{es} & \textsc{ru} & \textsc{zh} & \textsc{ja} & \textsc{th} & \textsc{te} & \textsc{bn} & \textsc{sw} \\
    \midrule
    Lang, freq. (\%) & - & - & - & 78.0 & 3.5 & 3.3 & 2.1 & 0.53 & 0.40 & 0.38 & 0.04 & 0.02 & 0.006 & 0.005 \\
    \midrule
    \underline{\PaLM} \vspace{0.5mm} \\
    \multicolumn{4}{l|}{\smallbullet Exemplar token length (avg.)}  & 95 & 108 & 119 & 105 & 113 & 118 & 118 & 193 & 199 & 173 & 130 \\
    \smallbullet \nativecot & \\
    \hspace{4mm} - 8B 6-shot & 4.0 & 4.1 & 3.1 & 6.4 & 6.8 & 4.4 & 2.4 & 2.8 & 4.0 & 4.4 & 3.2 & 3.6 & 3.2 & 2.4 \\
    \hspace{4mm} - 62B 6-shot & 20.0 & 22.7 & 13.2 & 30.4 & 24.0 & 24.0 & 26.0 & 22.8 & 24.8 & 14.8 & 18.0 & 11.6 & 13.6 & 9.6 \\
    \hspace{4mm} - 540B 1-shot & 38.9 & 39.7 & 34.8 & 50.8 & 42.8 & 44.8 & 44.8 & 41.2 & 34.8 & 29.6 & 40.0 & 38.0 & 34.0 & 27.2 \\
    \hspace{4mm} - 540B 2-shot & 43.7 & 44.0 & 39.8 & 57.2 & 47.2 & 43.2 & 50.4 & 44.4 & 44.4 & 34.4 & 47.2 & 38.0 & 40.8 & 33.2 \\
    \hspace{4mm} - 540B 4-shot & 45.1 & 45.5 & 41.0 & 58.8 & 44.8 & 49.6 & 47.6 & 46.4 & 46.4 & 38.4 & 46.4 & 41.2 & 44.4 & 32.0 \\
    \hspace{4mm} - 540B 6-shot & 48.1 & 47.9 & 44.9 & 62.4 & 49.2 & 46.4 & 56.8 & 48.4 & 46.8 & 40.0 & 52.8 & 45.6 & 46.0 & 35.2 \\
    \bottomrule
    \end{tabular}
    \caption{Detailed performances corresponding to \cref{fig:scaling_laws,fig:num_shots}.}
    \label{tab:appendix_table}
\end{table*}
\endgroup

We report the detailed numbers in our analysis (\cref{fig:scaling_laws,fig:num_shots}) in Table~\ref{tab:appendix_table}.
\clearpage
\section{The Chain-of-Thought Prompts Used in the Paper}
\begin{figure}[!ht]
\centering
\includegraphics[width=\linewidth]{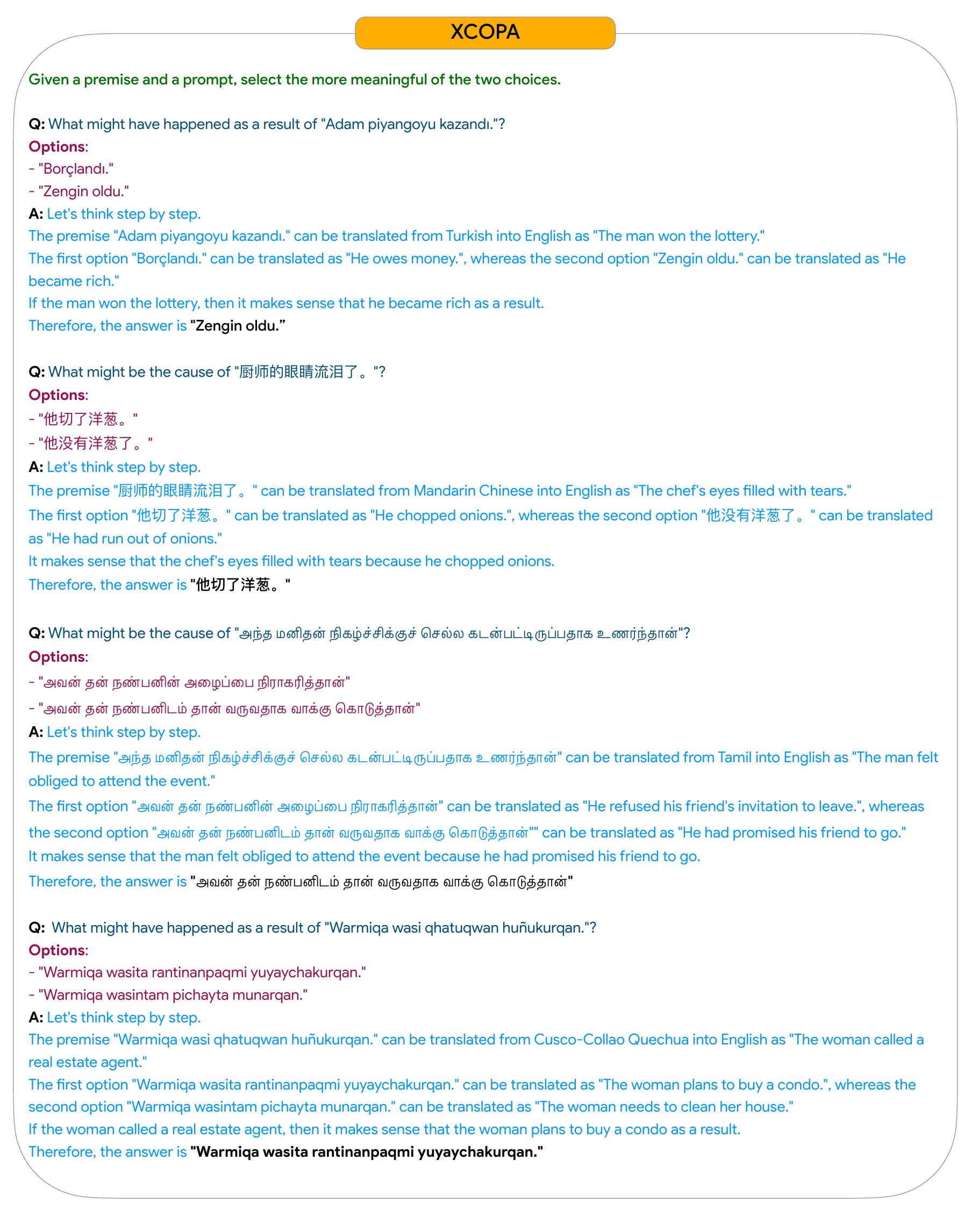}
\caption{The chain-of-thought prompt used in the \xcopa experiments. The four examples are randomly selected from the validation sets of Turkish (\textsc{tr}), Mandarin Chinese (\textsc{zh}), Tamil (\textsc{ta}), and Cusco-Collao Quechua (\textsc{qu}). The rationales are written by the authors, and the task description is taken directly from \citep{ponti-etal-2020-xcopa}. Under the direct prompting setup, the answers (\textbf{bolded}) are given directly and rationales are entirely omitted.}
\label{fig:pull-figure}
\end{figure}

\begin{figure}[!ht]
\centering
\includegraphics[width=\linewidth]{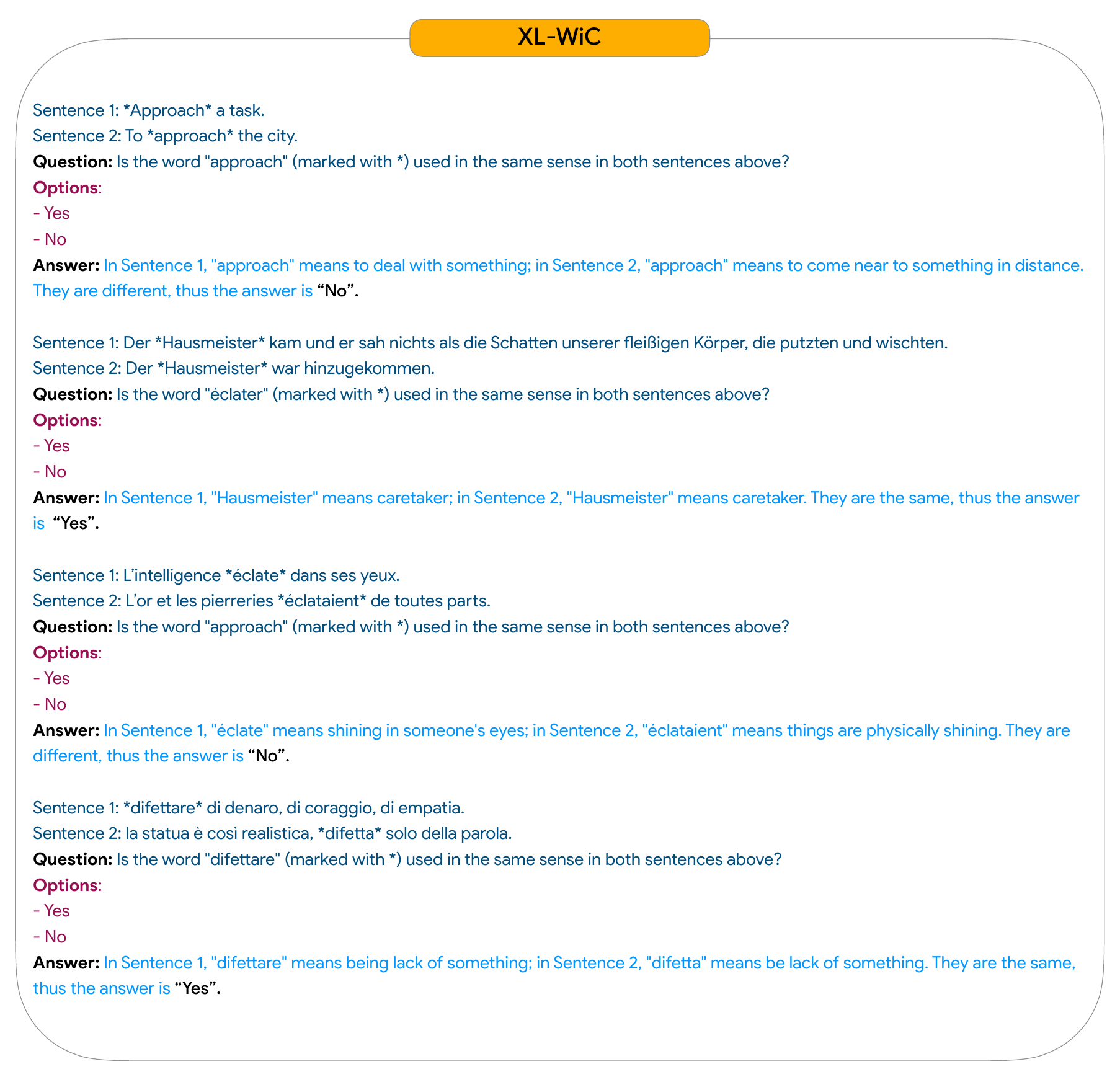}
\caption{The multilingual chain-of-thought prompt used in the \xlwic experiments.}
\label{fig:xl_wic_cot_v1}
\end{figure}

\begin{figure}[!ht]
\centering
\includegraphics[width=\linewidth]{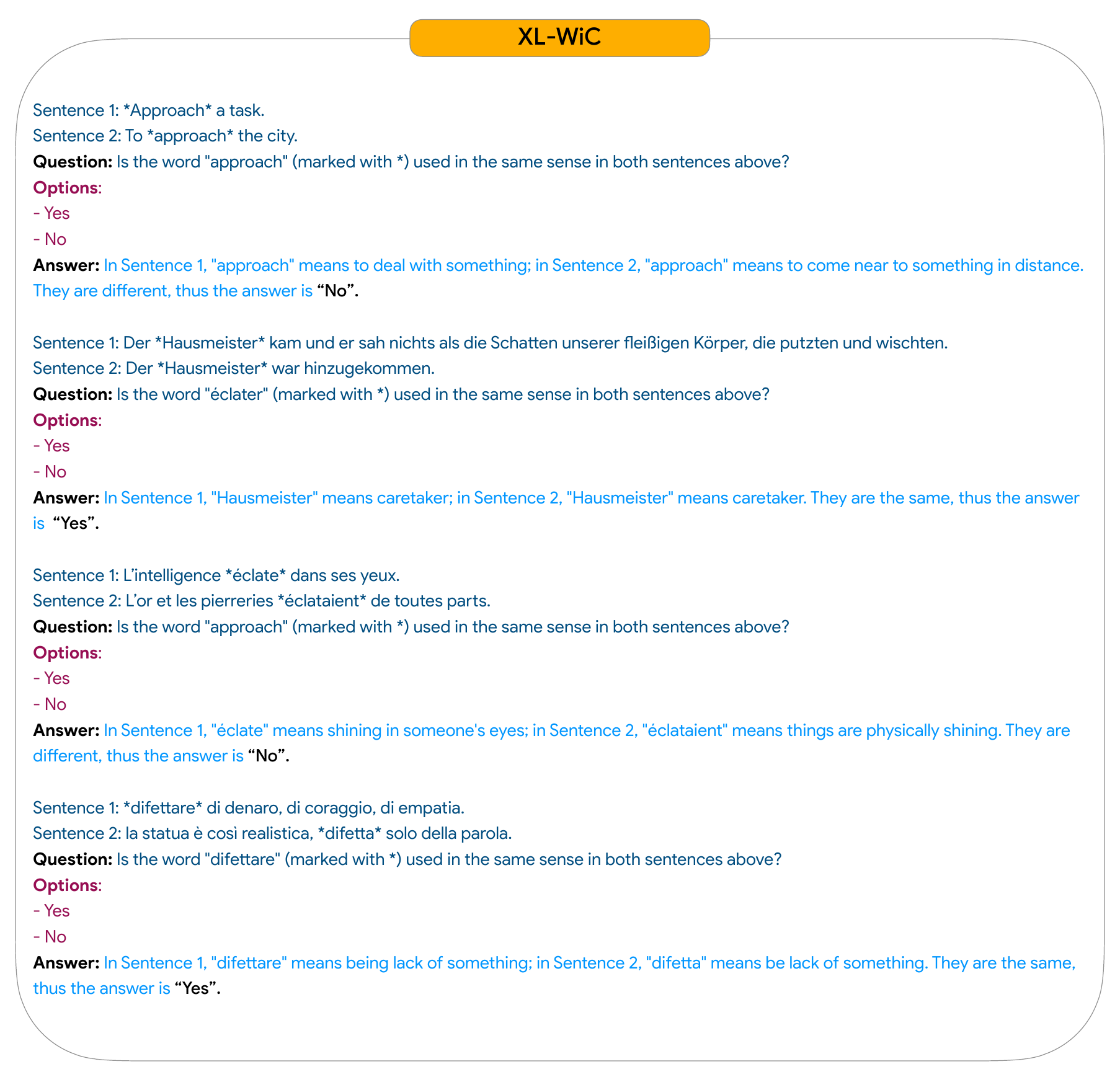}
\caption{The English-language chain-of-thought prompt used in the \xlwic experiments.}
\label{fig:xl_wic_cot_v2}
\end{figure}

In this section, we present the details of the chain-of-thought prompts used in our paper for the \xcopa (\cref{fig:pull-figure}) and the \xlwic (\cref{fig:xl_wic_cot_v1,fig:xl_wic_cot_v2}) tasks. 

\clearpage

\clearpage

\end{document}